\documentclass{article}
 \usepackage[dblblindworkshop, final, nonatbib]{neurips_2025}
\workshoptitle{Frontiers in Probabilistic Inference: Sampling Meets Learning}

\usepackage[utf8]{inputenc} 
\usepackage{biblatex} 
\usepackage[T1]{fontenc}    
\usepackage{hyperref}       
\usepackage{url}            
\usepackage{booktabs}       
\usepackage{amsfonts}       
\usepackage{nicefrac}       
\usepackage{microtype}      
\usepackage{xcolor}         
\usepackage{graphicx}
\usepackage{subcaption}     
\usepackage{algorithm}
\usepackage{algpseudocode}
\usepackage{amsmath}
\usepackage{amssymb}
\usepackage{amsthm}

\newtheorem{theorem}{Theorem}[section]
\newtheorem{lemma}[theorem]{Lemma}
\DeclareMathOperator*{\argmin}{arg\,min}

\title{Counterdiabatic Hamiltonian Monte Carlo}

\author{%
  Reuben Cohn-Gordon \\
  University of California, Berkeley \\
  \texttt{reubenharry@gmail.com} \\
  \And Uroš Seljak \\
  University of California, Berkeley
  \And
  Dries Sels \\
  New York University \\
}

\begin{document}

\maketitle

\begin{abstract}
  Hamiltonian Monte Carlo (HMC) is a state of the art method for sampling from distributions with differentiable densities, but can converge slowly when applied to challenging multimodal problems. Running HMC with a time varying Hamiltonian, in order to interpolate from an initial tractable distribution to the target of interest, can address this problem. In conjunction with a weighting scheme to eliminate bias, this can be viewed as a special case of Sequential Monte Carlo (SMC) sampling \cite{doucet2001introduction}. However, this approach can be inefficient, since it requires slow change between the initial and final distribution. Inspired by \cite{sels2017minimizing}, where a learned \emph{counterdiabatic} term added to the Hamiltonian allows for efficient quantum state preparation, we propose \emph{Counterdiabatic Hamiltonian Monte Carlo} (CHMC), which can be viewed as an SMC sampler with a more efficient kernel. We establish its relationship to recent proposals for accelerating gradient-based sampling with learned drift terms, and demonstrate on simple benchmark problems.
\end{abstract}

\section{Introduction}

Algorithms for sampling from probability distributions have many application in the physical sciences \cite{tuckerman2023statistical, von2011bayesian} and Bayesian statistics \cite{gamerman2006markov}. For sampling from  distributions with differentiable densities, \emph{Hamiltonian Monte Carlo} (HMC) \cite{betancourt2017conceptual} is a state of the art scheme, and in conjunction with the No-U-Turn criterion for trajectory length \cite{hoffman2014no}, is a widely used \emph{black box} method for inference \cite{gelman2015stan}.

Markov Chain Monte Carlo (MCMC), the family of algorithms to which HMC belongs, samples from a distribution with density $\pi$ by constructing an ergodic Markov kernel which has this target as its stationary distribution. However, the number of steps in the resulting Markov chain required for the law of the chain to converge to the target can be prohibitively large. A motivating example is a mixture of two Gaussians with means separated by a distance much larger than the respective variances; if the chain is initialized near one of the two modes, attaining stationarity requires moving through areas of low probability density.


\begin{figure}
    \centering
    \includegraphics[width=0.9\linewidth]{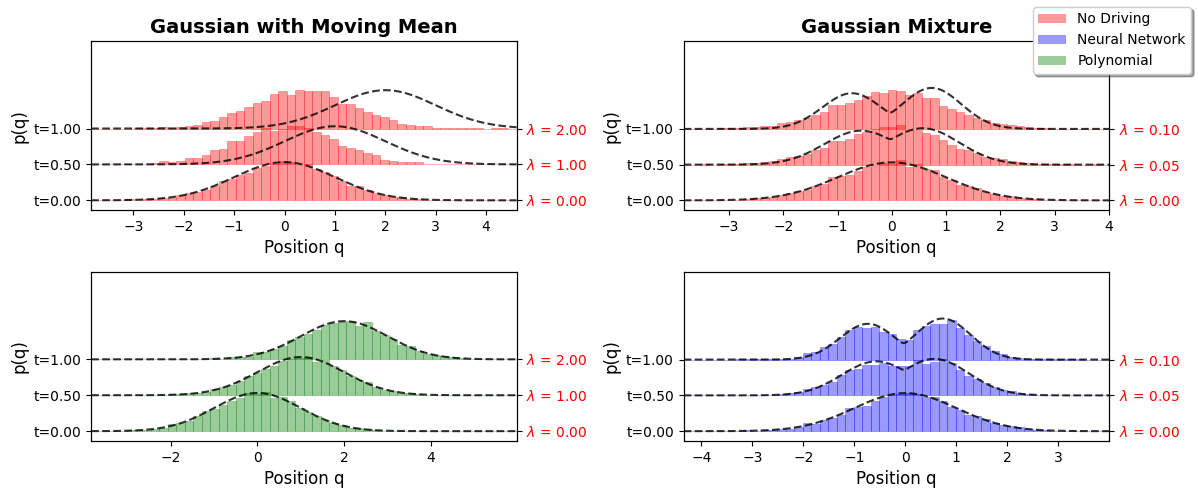}
    \caption{Illustration of lag induced by a time-varying Hamiltonian (top row), and the correction introduced by a learned \emph{counterdiabatic term} (bottom row). The lefthand column shows a Gaussian with time-varying mean, and the righthand column an interpolation between a Gaussian and a mixture of two Gaussians. In the simple case, parametrizing $A$ as a sum of polynomials with learned coefficients suffices, and in the more complex case, $A$ is parameterized by a neural network.
    }
    \label{fig:counterdiabatic_examples}
\end{figure}



An intuitive solution to this problem is to vary the target distribution over time, so that $\pi_0$ is easy to sample from, while $\pi_T$ is the desired target, and the kernel $M_t$ at time $t$ is an invariant kernel of $\pi_t$. Sequential Monte Carlo (SMC) samplers \cite{del2006sequential} do just this, running many chains (particles) in parallel, and using a weighting scheme to obtain unbiased samples from each $\pi_i$.

Hamiltonian Monte Carlo (HMC) is rooted in the connection between probabilistic inference and classical statistical physics, and constructs a Markov kernel $M$ by simulating Hamiltonian dynamics. These dynamics have the canonical (Boltzmann) distribution as a stationary distribution in an extended space of position $q$ and momentum $p$, and the Hamiltonian is chosen so that marginalizing over the momentum yields the target distribution $\pi$ on $q$.
In the context of HMC, moving through a series of kernels as above corresponds to evolving a physical system while changing its parameters, which is a central topic of statistical mechanics. From this perspective, varying the parameters too quickly results in \emph{lag}, or more physically, \emph{dissipation} \cite{vaikuntanathan2011escorted}, where the Hamiltonian flow cannot return the particles to equilibrium quickly enough as the system's parameters change. Two examples of this lag are shown in Figure~\ref{fig:counterdiabatic_examples} in red.


Several papers \cite{demirplak2003adiabatic, berry2009transitionless}, both in the quantum and classical setting, have noted that one can compensate for this lag by adding a \emph{counterdiabatic potential} to the Hamiltonian, which gives rise to a force that keeps the particles in the stationary distribution constantly. More recently, a scheme to approximate this term variationally has been proposed in a quantum setting \cite{sels2017minimizing}. 

\paragraph{Contributions} 
We adapt this variational counterdiabatic approach to the setting of sampling, to obtain a method for transitioning quickly between a simple distribution and the distribution of interest. This is shown in the green and blue plots of Figure~\ref{fig:counterdiabatic_examples}, where the counterdiabatic term accelerates the equilibration, so that a subsequent SMC reweighting scheme has less variance.
We note that our approach is closely related to the proposals of \cite{albergo2024nets, vargas2023transport, arbel2021annealed}, but in a Hamiltonian setting (see section \ref{sec:related}).






\section{Technical Background}

\paragraph{Hamiltonian dynamics}

Suppose we have a physical system given by a $d$-dimensional vector $q \in \mathbb{R}^d$, which describes, for example, the position of a particle. Hamilton's equations of motion are a 1st order ODE defined on $z=(q,p) \in \mathbb{R}^{2d}$, where $p \in \mathbb{R}^d$ is a second variable called the momentum. The ODE is defined in terms of a real-valued \emph{Hamiltonian} function $H(q,p)$.

\vspace{-0.4cm}
\begin{equation}
\dot{q} = \nabla_p H(q,p) ;\quad \quad
\dot{p} = -\nabla_q H(q,p)
\end{equation}


We refer to the solution to the ODE as the \emph{flow} $\phi_H$, so that $\phi_H(t)(q_0,p_0)$ is the state of the system at time $t$ given that the initial state is $z_0 = (q_0,p_0)$.

\paragraph{Poisson bracket} Given two real-valued functions of $(q,p)$, $f$ and $g$, the Poisson bracket $\{f,g\} \equiv \nabla_q f \cdot \nabla_p g - \nabla_p f \cdot \nabla_q g$. The bracket allows a convenient description of Hamiltonian dynamics. In particular, the total derivative $\frac{df(q(t), p(t))}{dt}$ can be written as $\{f, H\} = \nabla_q f \cdot \nabla_p H - \nabla_p f \cdot \nabla_q H = \nabla_q f \cdot \dot q + \nabla_p f \cdot \dot p =  \frac{df}{dt}$. We note that the Poisson bracket is antisymmetric, i.e. $\{A, B\} = -\{B, A\}$, and that if $f(q,p) = h(g(q,p))$ for some $h$, then $\{f, g\} =0$.

\paragraph{Canonical distribution} The distribution $\rho_H(q,p) \equiv (e^{-H(q,p)})/Z$, with normalization $Z=\int e^{-H(q,p)}dqdp$, is known as the \emph{canonical distribution}. When $H$ has no dependence on time except through $q$ and $p$, then $\frac{d\rho_H}{dt} = \{\rho_H, H\} = 0$, with the last equality following by an above property of the Poisson bracket.
We denote by $\phi^*_H$ the pushforward of the flow on a distribution, so that $\forall t, \phi^*_H(t)(\rho_H) = \rho_H$ is the statement that the canonical distribution is stationarity under the dynamics.





\paragraph{Hamiltonian Monte Carlo} The principle of Hamiltonian Monte Carlo is to use a discretization of Hamiltonian dynamics to produce a kernel which has a stationary distribution $\rho_H$ for which the marginal on $q$, i.e. $\pi(q) \propto \int \rho_H(q,p) dp$, is the distribution of interest. This can be achieved by choosing $H = -\log\pi(q) + \frac{1}{2}p^2 \equiv V(q) + T(p)$, which only requires knowing the density up to the normalization constant.
The Metropolis-Hastings algorithm \cite{chib1995understanding} is used to correct for bias coming from the discretization of the Hamiltonian dynamics, and momentum can be refreshed every $n$ steps, or partially refreshed every step \cite{cheng2018underdamped}, in order to ensure ergodicity.

\subsection{Counterdiabatic driving}


Consider $H_{\lambda(t)}$, a Hamiltonian which depends on a parameter $\lambda$ which varies with time. We can now consider the flow $\phi_{H_{\lambda}}$ that is the solution to Hamilton's equations. Crucially, we note that $\phi^*_{H_{\lambda}}(t')\rho_{H_{\lambda(t)}} \neq \rho_{H_{\lambda(t+t')}}$. In other words, the flow of the time-varying Hamiltonian does not preserve the time-varying distribution. 
To see why it fails, we inspect the total derivative:

\vspace{-0.5cm}

\begin{equation}
\frac{d}{dt}\rho_{H_\lambda} = \{\rho_{H_\lambda}, H_\lambda\} + \dot\lambda\partial_\lambda \rho_{H_\lambda}
\\
= \dot\lambda\partial_\lambda H\rho_{H_\lambda} - \dot\lambda\frac{\partial \log Z_\lambda}{\partial \lambda}\rho_{H_\lambda} = (-\partial_\lambda H_\lambda + \mathbb{E}[\partial_\lambda H_\lambda]) \dot\lambda\rho_{H_\lambda}
\end{equation}


where the expectation is over $\rho_{H_\lambda}$.
\cite{demirplak2003adiabatic}, \cite{berry2009transitionless} and \cite{vaikuntanathan2011escorted, guery2023driving} note
that one can express the lag in terms of a function $A_\lambda$ of $(q,p)$, known as the \emph{counterdiabatic gauge potential}, with the property that at any $\lambda$, 

\begin{equation}
\label{eq:prop}
\{A_\lambda, H_\lambda\} = \partial_\lambda H_\lambda - \mathbb{E}[\partial_\lambda H_\lambda]
\end{equation}

We then have that $\frac{d}{dt}\rho_{H_\lambda} = \dot \lambda\partial_\lambda\rho_{H_{\lambda}} = \dot\lambda\{H_\lambda, A_\lambda\}\rho_{H_\lambda} = \{\rho_{H_\lambda}, -\dot\lambda A_\lambda\}$.
This is the total derivative that would be generated by a Hamiltonian $-\dot\lambda A_\lambda$. This demonstrates a key result: if $A_\lambda$ exists, and if we choose as our Hamiltonian $\mathrm{HC}_{\lambda(t)} \equiv H_{\lambda(t)} + \dot\lambda(t) A_{\lambda(t)}$, we have that $\phi^*_{\mathrm{HC}_\lambda}(t')\rho_{H_{\lambda(t)}} = \rho_{H_{\lambda(t+t')}}$. Thus, by using $\mathrm{HC}_\lambda$ as our Hamiltonian, we obtain perfect transport to each $\rho_{H_{\lambda(t)}}$, including $\rho_{H_{\lambda(T)}}$. Note that Hamiltonian flow preserves volume and consequently entropy, so this can only transport between distributions with the same entropy. In line with Hamiltonian Monte Carlo, one solution is to introduce stochasticity in the dynamics of the momentum (see section \ref{sec:chmc}).


A simple example of a Gaussian with a time-varying mean is given in Appendix \ref{app:example}, where the counterdiabatic term can be derived analytically. In general however, it is necessary to learn $A_\lambda$.


\subsection{Learning the counterdiabatic gauge potential} 

\begin{lemma}
Let $\mathcal{L}_H(A) = \mathbb{E}[\left\|\{A, H\} - \partial_\lambda H\right\|^2]$, where the expectation is taken over $\rho_H$. Then equation \ref{eq:prop} is satisfied by

\vspace{-0.5cm}
$$A^* = \argmin_A \mathcal{L}_H(A)$$

\end{lemma}

This is shown in \cite{sels2017minimizing} in the quantum setting. We provide a similar proof in the present case in appendix \ref{app:variational}.

\section{Counterdiabatic Hamiltonian Monte Carlo (CHMC)} \label{sec:chmc}

\begin{algorithm}[t]
    \caption{Fit adiabatic gauge potential $A_\phi$ at a given $\lambda$}
    \label{alg:fit-agp}
    \begin{algorithmic}[1]
    \Require Real-valued function $A_\phi(q,p)$, optimizer (e.g. Adam)\;, schedule $\lambda_i$, population of particles $\{q_i,p_i\}_{i=1}^N$ and weights $w_{i=1}^N$, and $T$ iterations.
    \For{$t=1,\dots,T$}
      \State Compute $l  \gets \frac{1}{N}\sum_{i=1}^N w_i \left(\{A_\phi,H_\lambda\}(q_i,p_i) - \partial_\lambda V_\lambda(q_i)\right)^2$
      \State Update $\phi$ using optimizer with loss $l$
    \EndFor
    \State \Return $\phi$
    \end{algorithmic}
    \end{algorithm}

\begin{algorithm}[t]
    \caption{Counterdiabatic HMC (CHMC) with population and weighting}
    \label{alg:cd-hmc}
    \begin{algorithmic}[1]
    \Require Population $\{q_i^{(0)}\}_{i=1}^N$ from initial distribution\;; potential $V_\lambda(q) = -\nabla \log \pi_{\lambda}$ and gradients\;; step size $\varepsilon$\;; momentum refresh rate $n$\;; schedule $\{\lambda_k\}_{k=1}^L$ with $\lambda_1=0$ (easy) and $\lambda_L=1$ (target)\;.
    \State Initialize weights $w_i^{(0)} = 1/N$ for all $i$
    \For{$k=1$ to $L-1$}
      \For{$i=1$ to $N$}
        \State Set $(q,p) \leftarrow (q_i^{(k-1)}, p_i^{(k-1)})$
        \State If $k \mod n=0$, sample $p \sim \mathcal{N}(0,I)$
        \State Update $A_{\phi}$ using the weighted population (Algorithm 1)
        \State Integrate equations of motion:
        \[
          q_{\text{half}} \leftarrow q + \tfrac{\varepsilon}{2}\Big(p + \dot{\lambda}_k\, \nabla_p A_\phi(q,p)\Big)
        \]
        \[
          p \leftarrow p - \varepsilon\Big(\nabla_q V_{\lambda_{k}}(q_{\text{half}}) + \dot{\lambda}_{k}\,\nabla_q A_\phi(q_{\text{half}},p)\Big)
        \]
        \[
          q \leftarrow q_{\text{half}} + \tfrac{\varepsilon}{2}\Big(p + \dot{\lambda}_k\, \nabla_p A_\phi(q_{\text{half}},p)\Big)
        \]
        \State Increment work $W$:
        \[
          W_i \leftarrow H_{\lambda_{k}}(q,p) - H_{\lambda_{k-1}}(q_i^{(k-1)}, p_i^{(k-1)})
        \]
        \State Update particle: $q_i^{(k)} \leftarrow q$
        \State Update weight: $w_i^{(k)} \leftarrow w_i^{(k-1)} \exp(-W_i)$
      \EndFor
      \State Normalize weights: $w_i^{(k)} \leftarrow w_i^{(k)} / \sum_j w_j^{(k)}$
      \State Optionally resample if effective sample size is low
    \EndFor
    \State \Return weighted population $\{(q_i^{(L)}, w_i^{(L)})\}_{i=1}^N$
    \end{algorithmic}
    \end{algorithm}

The fact that we can learn an approximation of $A_\lambda$ suggests an algorithm for sampling. We first choose a path of densities $\pi_{\lambda(t)}$, where $\pi_{\lambda(0)}$ is chosen to be straightforward to sample from, and where $\pi_{\lambda(T)}$ is the distribution of interest. This defines a time-varying Hamiltonian $H_{\lambda(t)}(p,q) =  \frac{1}{2}|p|^2 - \log \pi_{\lambda(t)}(q) \equiv T(p) + V_{\lambda(t)}(q)$, and in turn a path of densities on the full space, $\rho_{H_{\lambda(t)}}$.

We define $A^\phi_\lambda$ as a neural network or sum of polynomials parametrized by a vector $\phi$. 
We initialize a population of particles distributed according to $\rho_{H_{\lambda(0)}}$. Choosing a step size $\epsilon$, so that at each step $j$, $t=j\epsilon$, we learn a new $\phi$ by minimizing $\mathcal{L}$, taking the expectation with respect to the current population (Algorithm \ref{alg:fit-agp}). We use $\mathrm{HC}_\lambda = H_{\lambda(t)} + \dot\lambda(t) A_{\lambda(t)}$ as our Hamiltonian at a given time $t$, and approximate Hamiltonian dynamics by a splitting integrator \cite{leimkuhler2004simulating}. Since $A$ is not generally separable, this will not be symplectic.

We can, and should, compute weights for each particle after every step, to obtain a consistent sampler. We explain this in more detail in Section \ref{sec:smc}, as a special case of Sequential Monte Carlo sampling \cite{del2006sequential}, and derive an expression for the weight from the non-equilibrium fluctuation theorem. The full algorithm is summarized in Algorithm \ref{alg:cd-hmc}.   

\section{CHMC in terms of Sequential Monte Carlo} \label{sec:smc}

Sequential Monte Carlo \cite{murphy2023probabilistic} is a general framework for probabilistic inference of path measures based on Feynman-Kac processes \cite{moral2004feynman}. The \emph{SMC sampler} approach (SMCS) \cite{del2006sequential, dai2022invitation} applies SMC to the problem of sampling from a distribution $\pi$ on $\mathbb{R}^d$. 
SMCS requires a series of densities $\pi_j$, with $\pi_0$ an initial distribution which is easy to sample from and $\pi_J$ the distribution of interest. We refer to the unnormalized density as $\gamma_j$, so that $\pi_j(x) = \frac{\gamma_j(x)}{\int \gamma_j(x)dx}$.

Markov kernels $M_j$ and $L_j$ must also be defined; $L_j$ only play a conceptual role in the algorithm, but the $M_j$ must be implemented.
The algorithm initializes a population of particles $\{x^i_0\}_{i=1}^N$, and at each step $j$ propagates them by a kernel $M_j$ to obtain $\{x^i_{j+1}\}_{i=1}^N$, and weights each particle $x^i_{j+1}$ by:

\begin{equation}
\label{weight}
w^i_j = \frac{\gamma_j(x_j^i)L_{j-1}(x^i_{j-1}|x^i_j)}{\gamma_{j-1}(x_{j-1}^i)M_{j}(x^i_{j}|x^i_{j-1})}
\end{equation}



\subsection{SMCS with physical dynamics: using the nonequilibrium fluctuation theorem}

In the case of interest to us, the variable in question is $z=(q,p)$, and the series of distributions is $\rho_{H_{\lambda(j\epsilon)}}(z)$, for integer $j$ and $\epsilon$ the step size of the discretization.
Further, we are concerned with kernels defined by Hamiltonian dynamics, so that $M^H_j(z'|z) = \delta(z'-\phi_{H(j\epsilon)}(\epsilon)(z))$ and $L^H_t(z|z') = \delta(\tilde z'-\phi_{H((j-1)\epsilon)}(\epsilon)(\tilde z))$. Here, for $z=(q,p)$, $\tilde z \equiv (q,-p)$. 
We have written $H(t)$ to denote $H_{\lambda(t)}$, with the understanding that $t$ varies over the course of the flow. We first consider the case of exact Hamiltonian dynamics, and then discuss the effect of discretization.


In this case, the ratio of $M$ and $L$ can be calculated by the non-equilibrium fluctuation theorem of \cite{crooks1998nonequilibrium}, which we now review. The value of this perspective is that this theorem can be generalized to various settings \cite{ohkuma2007fluctuation}, including to account for counterdiabatic driving.
First, we recall that

$$
\frac{dH}{dt} = \frac{\partial H}{\partial t} + \nabla H \cdot \frac{d\mathbf{z}}{dt} \equiv dW + dQ
$$

In physical terms, $dW$ and $dQ$ are the work and heat respectively, and mathematically, should be viewed as differential one-forms that are not exact, i.e. are not the differential of a function. By contrast, their sum $dH$ is an exact one-form, namely the differential of the Hamiltonian. In particular, the integral $\Delta H = \int_\alpha dH$ over a trajectory $\alpha$ is independent of $\alpha$, but is composed of two parts, $W[\alpha] = \int_\alpha dW$ and $Q[\alpha] = \int_\alpha dQ$ which are both dependent on $\alpha$, so that $\Delta H = W[\alpha] + Q[\alpha]$.

The fluctuation theorem states that

$$
\frac{\rho_H(z)P_{z\rightarrow z'}}{\rho_H(z')P_{z\leftarrow z'}} = \exp({W - \Delta F})
$$

where $P_{z\rightarrow z'}$ is the density over a given trajectory under the SDE in question (here Hamilton's equations of motion) from $z$ to $z'$, and where $W$ is the integral of $dW = \frac{\partial H}{\partial t}$ over the trajectory, and $\Delta F = -\log\frac{Z_{\lambda(t)}}{Z_{\lambda(t-1)}}$, for $Z_{\lambda} \equiv \int \rho_{H_{\lambda}}(z)dz$.

In the case of Hamiltonian dynamics, the density of paths is precisely the delta function $\delta(z'- \phi_H(z))$, so a rearrangement gives us $w^i_j = e^{-W}$, where $W$ is calculated over the trajectories of $w^i$. Further, we have that

$$
\frac{M_j(z_j | z_{j-1})}{L_{j-1}(z_{j-1} | z_j)} = e^W\frac{Z_j}{Z_{j-1}}\frac{Z_{j-1}}{Z_j}e^{-\Delta H} = e^{W - \Delta H} = e^{-Q}
$$

where $Z_j = Z_{\lambda(j\epsilon)}$. Since Hamiltonian dynamics are measure-preserving, i.e. $Q=0$, we see immediately that when using an unadjusted HMC kernel in the SMC sampling algorithm, the correct weight is given by $\frac{\gamma_{j+1}}{\gamma_j} = e^{-\Delta H}$, in agreement with Appendix B of \cite{dai2022invitation}.

The more interesting case concerns kernels $M_j^{\mathrm{HC}_\lambda}$ arising from a counterdiabatic Hamiltonian $\mathrm{HC}_\lambda = H + \dot\lambda A$. In this case, \cite{zhong2024time} derives an analogous fluctuation theorem, but in terms of a different work, which we refer to as $W_c$, obtained by integrating $dW_c$ defined as follows:

$$
dW_c = \left(\frac{\partial H}{\partial t} + \nabla H\cdot v_A - \nabla \cdot v_A\right)dt
$$

where $v_A$ is the drift induced by $\dot\lambda A$, i.e. $v_A = (\nabla_q \dot\lambda A, -\nabla_p \dot\lambda A)$. We then see that the third term vanishes, since Hamiltonian flow is dissipationless. 
Then the physical intepretation of heat and work is preserved, in the sense that 

$$
\frac{dH}{dt} = dW_c + dQ
$$

where $dQ$ is as defined above, i.e. $dQ = \nabla H \cdot v_H$, with $v_H = (\nabla_q H, -\nabla_p H)$. The result is that the weight is the negative exponential of the change not in $HC$, but in the original $H$: $w \propto e^{-\Delta H}$.


\subsection{Discretization} The above discussion assumes exact Hamiltonian dynamics, but in practice, we must numerically integrate. It then remains to show that the same weight $e^{-\Delta H}$ applies. This can be shown by direct calculation (see e.g. Appendix 6.2 of \cite{albergo2024nets}) or indirectly, by viewing discretized Hamiltonian dynamics as the exact dynamics of a so-called shadow Hamiltonian $H_S$, so that the work ``performed'' by discretization can be understood as an instantaneous change from $H$ to $H_S$ before the leapfrog step, and a corresponding change back afterwards. \cite{sivak2013using} refer to the work from the change of $\lambda$ and from discretization as protocol work $W_P$ and shadow work $W_S$ respectively, with $\Delta H = W_S + W_P + Q$. Because the fluctuation theorem holds in the context of total work $W = W_S + W_P$, and because Hamiltonian dynamics have $Q=0$, we have that $W = \Delta H$ and so $w \propto e^{-\Delta H}$.
The counterdiabatic case can be approached similarly. 



\begin{figure}
    \centering
    \includegraphics[width=0.9\linewidth]{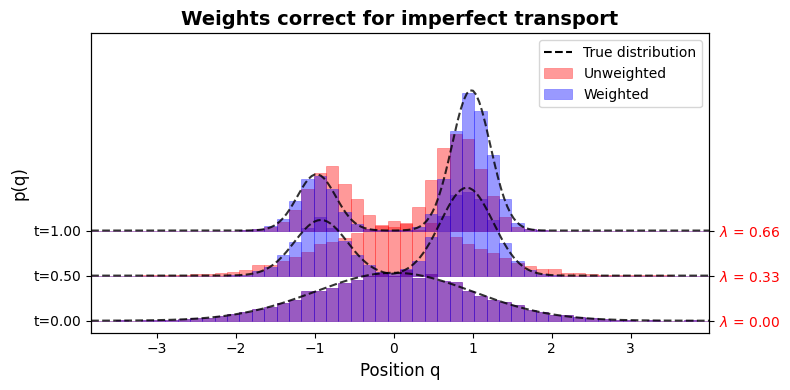}
    \caption{Here, the path of distributions is a geometric path between a standard normal and a mixture of two }
    \label{fig:weighting}
\end{figure}

As a demonstration of the weighting scheme working in practice, we show in Figure \ref{fig:weighting} a SMC scheme using the counterdiabatic kernel. 

\section{Related work}
\label{sec:related}

From the perspective of stochastic optimal control \cite{fleming2012deterministic} and optimal transport \cite{peyre2019computational}, the problem at hand is to design an ODE or SDE
that induces a desired trajectory in the space of densities, from some $\pi_0$ to the target $\pi_T$.
In this setting, Controlled Monte Carlo Diffusions \cite{vargas2023transport}, the Liouville Flow Importance Sampler \cite{tian2024liouville}, Path Guided Particle Sampling \cite{fan2024path}, the Path-Integral Sampler \cite{zhang2021path}, the Adjoint sampler \cite{havens2025adjoint} and Non-equilibrium transport sampling \cite{albergo2024nets} all learn a drift term of a differential equation, parametrized by a neural network.

Several of these approaches use weights to correct for remaining bias \cite{albergo2024nets, vargas2023transport}. \cite{albergo2024nets} in particular presents a similar scheme to ours, but with overdamped Langevin dynamics, rather than Hamiltonian dynamics,  with different loss functions, and with weights applied non-incrementally and not used in training.
With weighting, these approaches can be understood in terms of Sequential Monte Carlo sampling \cite{dai2022invitation}, as described in Appendix \ref{app:smc}, and in that setting, \cite{heng2020controlled} and \cite{bernton2019schr} have both considered learning kernels $M_t$ instead of setting $M_t$ as the invariant kernel of each successive $\pi_t$.


Finally, adding a term to Hamiltonian dynamics to preserve stationarity through a series of intermediate distributions has been proposed several times in a quantum setting, as \emph{engineered swift equilibration} \cite{martinez2016engineered} and counterdiabatic driving \cite{demirplak2003adiabatic}, where the corresponding problem with a quickly changing Hamiltonian is failure to preserve eigenstates. The core ideas carry over to a classical setting \cite{patra2017shortcuts}, often under the name \emph{shortcuts to adiabaticity}, or \emph{escorted free energy} (EFE) \cite{vaikuntanathan2011escorted}. We recommend \cite{guery2019shortcuts} for an overview. 
\cite{sels2017minimizing} proposes \emph{learning} a driving term in a quantum context, but notes that the approach is transferable to a classical setting, which is the foundation for our approach.
\vspace{-0.3cm}
\paragraph{Where CHMC differs from related work} The work summarized above focuses either on overdamped dynamics, or Hamiltonian dynamics without learning. Since Hamiltonian Monte Carlo is the gold standard approach for sampling from differentiable densities, because of the geometric properties of Hamiltonian dynamics \cite{betancourt2017geometric}, it is natural that schemes for accelerating sampling in multimodal targets should focus on it. In this Hamiltonian setting, the learned term takes on an geometric significance, as a connection on a fiber bundle \cite{kolodrubetz2017geometry}. The use of this geometry, and in particular the curvature, to determine optimal paths through the space of distributions, is a topic for future work.

\vspace{-0.4cm}
\section{Evaluation}
\vspace{-0.2cm}

Here, we show the difference, on several simple problems, between the final particles that result from using CHMC and standard HMC, in order to highlight how counterdiabatic driving accelerates transport. We note that the goal is not to obtain unbiased samples from the target, but rather to improve the performance of subsequent SMC-style reweighting. We measure the standardized squared error $b^2(f) = \frac{(\hat f - \mathbb{E}[f])^2}{\mathrm{Var}[f]}$, where $\hat f$ is the Monte Carlo estimate $\sum_i^Nf(q_i)$ of the final population of samples $\{q_i\}$ and $f(q) \equiv q^2$.
We consider 3 problems: a 1D Gaussian with a moving mean, a 1D Gaussian with a changing variance, and an interpolation between a 1D Gaussian. See Appendix \ref{app:implementation} for details.

\vspace{-0.2cm}

\begin{table}[h]
\centering
\label{tab:second_moment_results}
\begin{tabular}{lccccc}
\toprule
System & True $E[x^2]$ & Naive HMC & CHMC \\
\midrule
Gaussian Moving Mean & 2.0 & 1.12 & 2.1  \\
Gaussian Annealing & 0.1 & 86.5 & 0.65  \\
Double Well & 7.34 & 2.06 & 4.22 \\
\bottomrule
\end{tabular}
\label{table:results}
\end{table}

The takeaway is that when running at a fast schedule, standard HMC results in poor transport, because more than a single step of the kernel at each time is required to regain stationarity, but that using the counterdiabatic Hamiltonian results in better transport. A more complete evaluation, which includes the use of weights, a more powerful neural network, and higher dimensional benchmark problems in the spirit of \cite{albergo2024nets}, is beyond the present scope, but a topic for future work. 




\section{Conclusion}

We have presented an application of learned counterdiabatic driving to the problem of sampling. This is closely related to the learned transport approach of \cite{albergo2024nets} among other recent papers, but introduces a new physically inspired perspective. In particular, the extensive work in the physics literature on the choice of parametrization for $A_\lambda$ and an optimal choice of $\lambda(t)$, and in applied statistics for the choice of $\epsilon$ and a mass matrix in HMC, suggest avenues for future work.

\printbibliography

\appendix

\section{Counterdiabatic HMC as a Sequential Monte Carlo (SMC) sampler}
\label{app:smc}







\section{Example of simple counterdiabatic driving term}
\label{app:example}

Suppose our time varying density $\pi_{\lambda(t)}$ is defined as a 1-dimensional Gaussian with unit variance and a time varying mean $\lambda(t)$, so that $\lambda(0)=0$ and $\lambda(T)=1$.

In this case, $A$ is known, and is equal to $p$. To see this, calculate $\{p,H\} = -\frac{\partial H_\lambda}{\partial q} = -\frac{\partial V_\lambda}{\partial q}$, while $\partial_\lambda H_\lambda = -\frac{\partial V}{\partial q}$ also.

Then we have $\{A_\lambda, H_\lambda\} = \partial_\lambda H_\lambda$, which implies that $\mathbb{E}[\partial_\lambda H_\lambda] = \mathbb{E}[\{A,H\}] = \int \{A,H\}\rho_H = \int \{H, \rho_H\}A = 0$, by a standard property of the Poisson bracket.

A term $p$ in the Hamiltonian generates motion towards the moving mean, so that probability mass is transported exactly along with the changing mean under the flow of the Hamiltonian.

It is also useful to consider the weight $w$ in this setting, which as per Appendix \ref{app:smc} is defined as $e^{-\int dW_c}$, for

$$
dW_c = \left(\frac{\partial H}{\partial t} + \nabla H\cdot v_A \right)dt
$$

Since the transport is perfect, we know that the weights should be uniform, so the two terms in $dW_c$ must cancel. Indeed, inspection shows that $\frac{\partial H}{\partial t} = -\dot\lambda(q-\lambda)$, while $\nabla H \cdot v_A = \dot\lambda(q-\lambda)$, giving the desired result. 





\section{Implementation of Counterdiabatic Hamiltonian Monte Carlo}
\label{app:implementation}

Here we review a number of details regarding the implementation of CHMC. Our implementation is in Jax, and for the learning of $A$ we use the Adam optimizer. We use a 2 hidden layer feedforward neural network with ReLu activations, and hidden layers of 32 and 64. As an alternative, we implement a polynomial approximation, i.e. a sum of all polynomials in $q$ and $p$ up to a given order (our default is $5$), where $\phi$ supplies the coefficients of each polynomial term. The latter is computationally cheaper to run and in low dimensions yields comparable results.

The code is available at \url{https://github.com/reubenharry/counterdiabatic-hmc/tree/working_branch}.

\paragraph{Choice of step size} As in standard HMC, the choice of $\epsilon$ is a trade-off between the accuracy of the dynamics and the efficiency of the algorithm. We conjecture that adaptive tuning, as in tuning schemes for HMC step size \cite{gelman2015stan} will prove useful. An advantage of the Hamiltonian setting is that large energy error can be used as a proxy for excessive step size.


\paragraph{Numerical Integration} A key complication is that $A_\lambda$ is not separable into a term depending on $p$ and a term depending on $q$. This means that the Velocity Verlet integrator is not symplectic. In the current work, we overlook this problem and make sure to choose $\epsilon$ reasonably small.

\paragraph{Momentum refreshing} In standard HMC, it is necessary to refresh momentum, either totally every $n_l$ steps, or partially every step. The latter corresponds to a discretization of underdamped Langevin dynamics. In the case of CHMC, we find that momentum resampling is valuable. We use the strategy of refreshing every $n_l$ steps, since the Langevin strategy results in dissipative dynamics which complicates the weight calculation. We fix $n_l=2$, but note that performance is likely to be sensitive to this value, as in normal HMC, so we view automatically determining $n_l$ as an important topic for future work.

For our experiments, we choose $\lambda(t)=0.5t$. For the Gaussian Annealing and the Gaussian Moving Mean systems, we use $\epsilon=2/3$ and $4$ total timesteps, and for the Double Well, we use $\epsilon=0.2$ and $10$ timesteps. We use $1000$ particles.

\subsection{System Potentials}

We evaluate our method on three one-dimensional systems with time-dependent potentials $V_{\lambda}(q)$.

\subsubsection{Gaussian Moving Mean}
The potential interpolates from a standard Gaussian to a Gaussian with shifted mean:
\begin{equation}
V_\lambda(q) = \frac{1}{2}(q - \lambda)^2
\end{equation}
This system evolves from $\mathcal{N}(0, 1)$ at $\lambda = 0$ to $\mathcal{N}(1, 1)$ at $\lambda = 1$.

\subsubsection{Gaussian Annealing}
The potential interpolates from a standard Gaussian to a Gaussian with reduced variance:
\begin{equation}
V_\lambda(q) = \frac{1}{2}k(\lambda) q^2
\end{equation}
where $k(\lambda) = 1 + 9\lambda$ interpolates from 1 to 10. This system evolves from $\mathcal{N}(0, 1)$ at $\lambda = 0$ to $\mathcal{N}(0, 0.1)$ at $\lambda = 1$.

\subsubsection{Double Well}
The potential interpolates from a harmonic oscillator to a double-well potential:
\begin{equation}
V_\lambda(q) = (1-\lambda)\frac{1}{2}q^2 + \lambda\left((q^2 - 3)^2\right)
\end{equation}
This system evolves from a single well centered at $q = 0$ to a double well with minima at $q = \pm\sqrt{3}$.

\subsection{Kinetic Energy}
All systems use the standard kinetic energy:
\begin{equation}
T(p) = \frac{p^2}{2m}
\end{equation}
with $m = 1$.

\section{Learning $A$}
\label{app:variational}

To learn $A$ satisfying property \ref{eq:prop}, it suffices to minimize $\mathbb{E}_{z \sim \rho_{H_{\lambda(t)}}}[|G(z)|^2]$, where

$$
G(z) = \{A, H\}(z) - \partial_\lambda H(z) + \mathbb{E}[ \partial_\lambda H(z)]
$$

Multiplying out, we find $\mathbb{E}[|G(z)|^2] = \mathbb{E}[\{A, H\}^2 + (\partial_\lambda H)^2 + \mathbb{E}[\partial_\lambda H]^2 - 2\{A, H\}(\partial_\lambda H) + 2\{A, H\}\mathbb{E}[\partial_\lambda H] - 2(\partial_\lambda H)\mathbb{E}[\partial_\lambda H]] = \mathbb{E}[\{A, H\}^2 + (\partial_\lambda H)^2 - 2\{A, H\}(\partial_\lambda H) + 2\{A, H\}\mathbb{E}[\partial_\lambda H] - \mathbb{E}[ \partial_\lambda H]^2]$.

The final term is irrelevant since it is constant with respect to $A$, while the penultimate term can be shown to vanish, since $\mathbb{E}[\{A,H\}] = \int \{A,H\}\rho_H = \int \{H, \rho_H\}A = 0$, by standard properties of the Poisson bracket.


This means that $\mathbb{E}[G(z)] = \mathbb{E}[\{A, H\}^2 + (\partial_\lambda H)^2 - 2\{A, H\}(\partial_\lambda H)] = \mathbb{E}[|\{A, H\} - \partial_\lambda H|^2] \equiv \mathcal{L}_H(A)$.


This means that we can minimize the loss $\mathcal{L}_H(A)$ instead, which is simple to calculate.

\subsection{Choice of Gauge}

The Hamiltonian $\mathrm{HC}_\lambda$ is presented in the main text as $H_\lambda + \dot\lambda A_\lambda$, but we are in fact free to choose instead of the first term $H_\lambda$, any function $K$ of $(q,p)$ which has $\{K, \rho_H\} = 0$. This is because, for a perfectly learned $A$, with $\epsilon\to 0$, such a term generates a flow under which $\rho_H$ is stationary. In physical language this is known as a choice of \emph{gauge}.

However, in the relevant regime, where $\epsilon$ is finite and $A_\lambda$ is merely approximate, the question of the ideal gauge is open. Intuitively, using $H_\lambda$ should allow the particles to (partially) re-equilibrate to the current stationary distribution at each step, and this is our justification for this choice. We also note that using $H_\lambda$ is computationally cheap; although it requires the use of expensive gradients of the density $\pi_\lambda$, these are needed anyway in the fitting of $A_\lambda$, which happens subsequently.

\newpage

\end{document}